\documentclass[letterpaper]{article} % DO NOT CHANGE THIS
\usepackage{aaai2026}  % DO NOT CHANGE THIS
\usepackage{times}  % DO NOT CHANGE THIS
\usepackage{helvet}  % DO NOT CHANGE THIS
\usepackage{courier}  % DO NOT CHANGE THIS
\usepackage[hyphens]{url}  % DO NOT CHANGE THIS
\usepackage{graphicx} % DO NOT CHANGE THIS
\usepackage{natbib}  % DO NOT CHANGE THIS AND DO NOT ADD ANY OPTIONS TO IT
\usepackage{caption} % DO NOT CHANGE THIS AND DO NOT ADD ANY OPTIONS TO IT
\usepackage{algorithm}
\usepackage{algorithmic}

\usepackage{newfloat}
\usepackage{listings}
\DeclareCaptionStyle{ruled}{labelfont=normalfont,labelsep=colon,strut=off} % DO NOT CHANGE THIS
\floatstyle{ruled}
\newfloat{listing}{tb}{lst}{}
\floatname{listing}{Listing}
\usepackage{booktabs}
\usepackage{amssymb}
\usepackage{multirow}
\usepackage{marvosym}

\makeatletter
\renewcommand{\@makefntext}[1]{%
  \parindent 0em\noindent~#1% % 没有缩进，直接显示标记和内容
}
\makeatother

\title{Dereflection Any Image with Diffusion Priors and Diversified Data}
\author{
    Jichen Hu\textsuperscript{\rm 1}\footnotemark[1],
    Chen Yang\textsuperscript{\rm 1}\footnotemark[1],
    Zanwei Zhou\textsuperscript{\rm 1},
    Jiemin Fang\textsuperscript{\rm 2}\footnotemark[2],
    Qi Tian\textsuperscript{\rm 2},
    Wei Shen\textsuperscript{\rm 1}\textsuperscript{\Letter}\footnotemark[2]
}
\affiliations{
    \textsuperscript{\rm 1} {MoE Key Lab of Artificial Intelligence, AI Institute, School of Computer Science, Shanghai Jiao Tong University}
\\
\textsuperscript{\rm 2} {Huawei Inc.}\\
 \texttt{\small\{hujichen, sjtu19zzw, wei.shen\}@sjtu.edu.cn}\;\; 
    \texttt{\small \{chenyang.res, jaminfong\}@gmail.com} \;\;
    \texttt{\small tian.qi1@huawei.com}
}

\usepackage{bibentry}
\begin{document}

\maketitle

\begin{figure*}[!t]
  \centering
  \includegraphics[width=\textwidth]{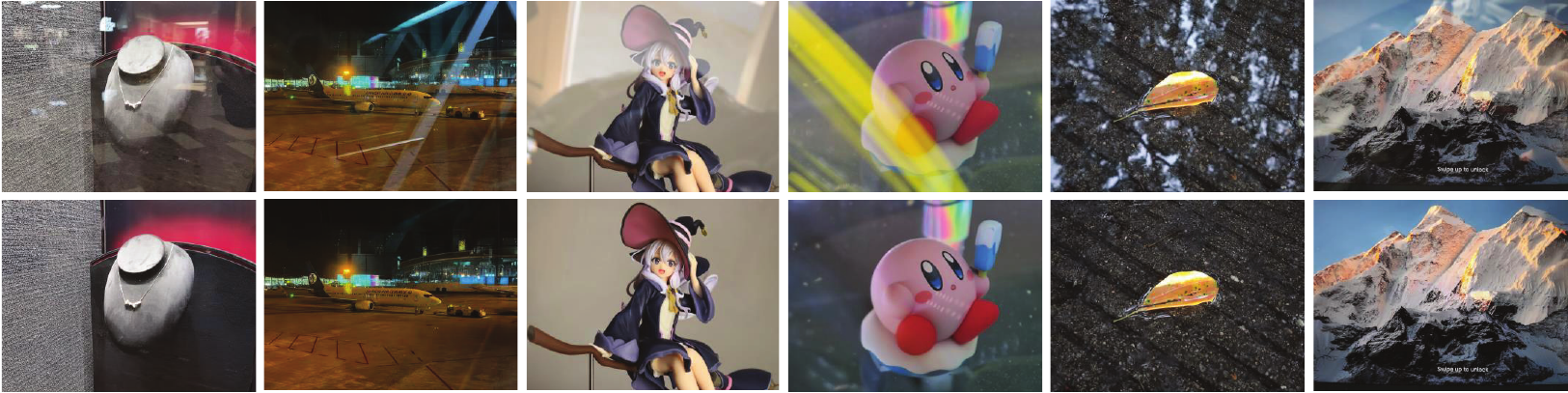}
  \captionsetup{type=figure} % 声明这是figure类型的标题
  \captionof{figure}{
    Our model demonstrates strong and general reflection removal capabilities.
    \textbf{Upper:} Original images with reflections. 
    \textbf{Bottom:} Results generated by our model.
    The scenarios include glass, plastic, water surfaces, \textit{etc.} 
  }
  \label{fig:teaser}
\end{figure*}

\renewcommand{\thefootnote}{\fnsymbol{footnote}} % 使用符号作为脚注标记
\footnotetext[1]{* Equal contribution.}
\footnotetext[2]{$\dagger$ Project lead.}
\footnotetext[3]{\Letter~Corresponding author.}

\renewcommand{\thefootnote}{\arabic{footnote}} % 恢复数字脚注
\setcounter{footnote}{0} % 重置脚注计数器

\begin{abstract}
Reflection removal of a single image remains a highly challenging task due to the complex entanglement between target scenes and unwanted reflections. Despite significant progress, existing methods are hindered by the scarcity of high-quality, diverse data and insufficient restoration priors, resulting in limited generalization across various real-world scenarios. 
In this paper, we propose Dereflection Any Image\footnote{``Dereflection'' may not be an accurate word, which is used to represent removing reflection for simplicity. This name is to respect \cite{kirillov2023segment, yang2024depth}.}, a comprehensive solution with an efficient data preparation pipeline and a generalizable model for robust reflection removal.
First, we introduce a dataset named Diverse Reflection Removal (DRR) created by randomly rotating reflective mediums in target scenes, enabling variation of reflection angles and intensities and setting a new benchmark in scale, quality, and diversity.
Second, we propose a diffusion-based framework with one-step diffusion for deterministic outputs and fast inference. To ensure stable learning, we design a three-stage progressive training strategy that includes reflection-invariant finetuning to encourage consistent outputs across varying reflection patterns that characterize our dataset.
Extensive experiments show that our method achieves SOTA performance on both common benchmarks and challenging in-the-wild images, showing superior generalization across diverse real-world scenes.
\end{abstract}

\begin{links}
    % \link{Code}{https://github.com/Abuuu122/Dereflection-Any-Image}
    \link{Project Page}{https://abuuu122.github.io/DAI.github.io}
\end{links}

\section{Introduction}
\label{sec:intro}
% Background and importance
Capturing images through glass or other reflective mediums often introduces unwanted reflections, which degrade the visibility of the underlying target scene, resulting in a mixed image with two layers superposed. These reflections significantly degrade both visual aesthetics and usages in downstream tasks~\cite{liu2020reflection, wan2021face}.
Developing strong reflection-removal (\textbf{dereflection} for short) methods is essential for practical applications.

Traditional methods~\cite{li2014single, shih2015reflection, wan2016depth} for reflection removal rely on empirical assumptions like reflections being blurred or exhibiting ghosting effects, which often fail in real-world scenarios. Learning-based methods~\cite{hong2024differ, hu2023single, hu2024single, song2023robust, zhong2024language, dong2021location, hu2021trash, zhu2023Weather} attempt to learn dereflection capabilities through paired data. 
However, such paired data is difficult to obtain in real-world settings, \textit{e.g.}, capturing museum exhibits with and without reflections typically requires physically removing the glass from display cases, which is often impractical.
Consequently, existing real-world datasets~\cite{wan2017benchmarking, zhang2018single, li2020single, zhu2023Weather} are limited in scale and diversity, failing to capture a wide range of reflection patterns encountered in practice. Moreover, synthetic datasets~\cite{wen2019single, kim2020single, hu2023single} often suffer from significant domain gaps, as their reflection patterns and image characteristics do not fully align with real-world scenarios. 

\begin{figure*}[t]
\captionsetup{type=figure}
\centering
\resizebox{\linewidth}{!}{
\includegraphics[width=1.0\textwidth]{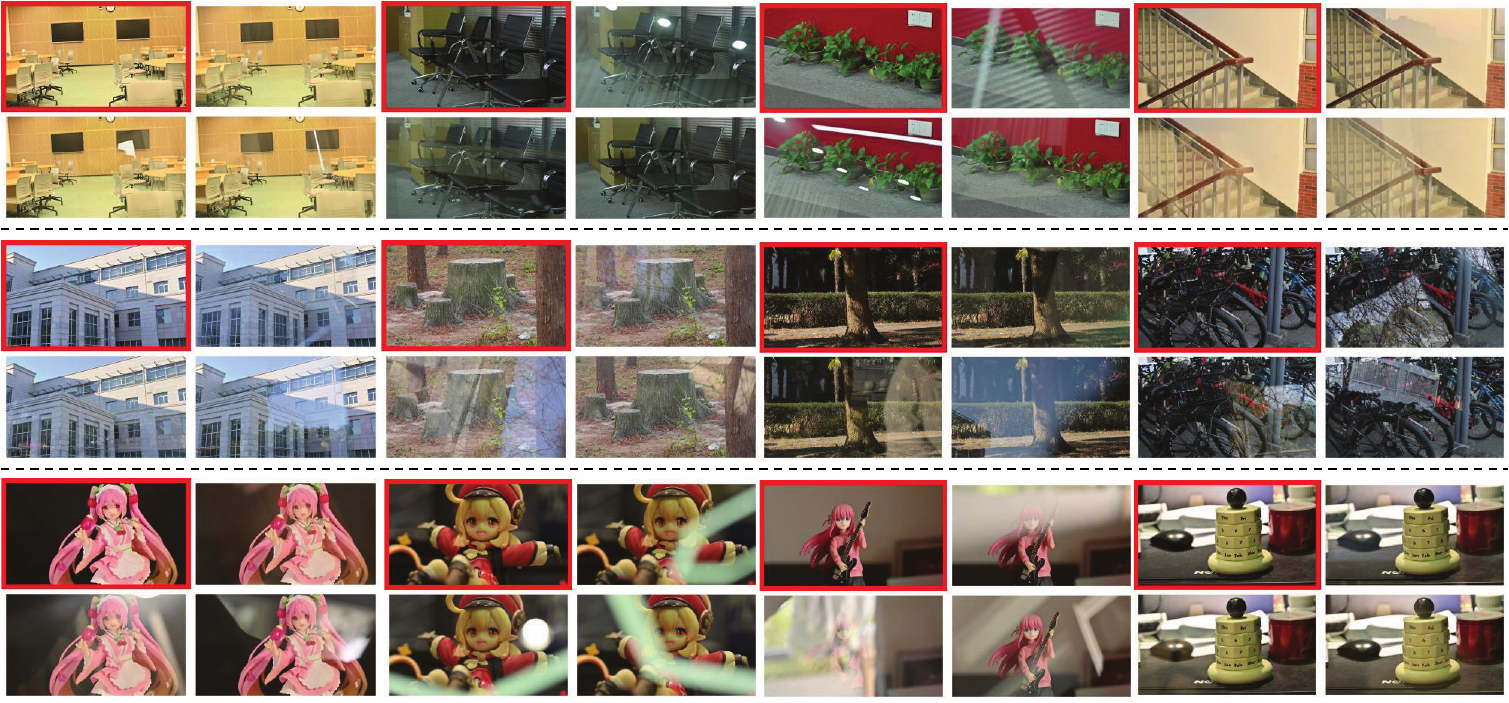}}
\caption {Our dataset contains a diverse collection of scenes, each accompanied by multiple reflection images. As illustrated in the figure, the ground truth transmission layer is highlighted in red boxes, while the remaining images represent various mixed images. The dataset demonstrates remarkable diversity, encompassing \textbf{indoor}, \textbf{outdoor}, and \textbf{object-centric scenes}. All image pairs maintain high resolution with rich textual details. (Best viewed on screen.) }
\label{fig: dataset}
\end{figure*}

To address these limitations, we propose an efficient data collection pipeline and introduce Diverse Reflection Removal (DRR), a 4K dataset with diverse reflection patterns. Specifically, for each view of the capturing device, reflective mediums are randomly rotated with the target scene. Then we capture mixed videos and decompose them into frames to form data pairs. This approach allows us to flexibly vary reflection angles, intensities, and scene diversity, significantly enhancing the realism and variety of reflection patterns compared to existing datasets~\cite{wan2017benchmarking,zhang2018single,li2020single,zhu2023Weather}, as shown in Table~\ref{tab:dataset_comparison} and Fig.~\ref{fig: dataset}.
We also construct synthetic pairs to strengthen the data sufficiency, which are filtered with the CLIP score~\cite{radford2021learning} to guarantee the data realism. Besides extending the diversity of training data, we propose to leverage the generalization capabilities of diffusion models, which excel in image-to-image translation tasks even tuned with finite data~\cite{xia2023diffir, tumanyan2023plug}. Our model integrates ControlNet~\cite{zhang2023adding} to use the mixed image as a conditioning signal.
To ensure deterministic outputs for reliable transmission layer recovery, we build our generative prior-based model on one-step diffusion~\cite{ye2024stablenormal, xu2024matters}, enabling both deterministic results and fast inference.
To ensure stable learning of translation from complex mixed images to dereflectioned ones, we design a three-stage progressive training strategy. Initially, we employ foundation training with image pairs to get basic performance, We further exploit the unique characteristics of our dataset through a reflection-invariant finetuning strategy. Since our dataset contains identical transmission scenes with varying reflection patterns, we train the model to produce consistent outputs despite these variations, enhancing generalization by focusing on invariant properties of transmission scenes rather than variable reflection characteristics. Finally, a cross-latent decoder is trained to mitigate blurriness and preserve details. The new dataset, code and model checkpoints will be released. 

Our key contributions are summarized as follows:
\begin{itemize}
    \item We introduce an efficient pipeline for data collection, and present DRR, a high-quality dataset featuring diverse reflections with varying angles, fostering future advancements in the field.
    \item We design a novel diffusion-based framework with a progressive training strategy ensuring both stable optimization and strong generalization across diverse reflection types, as shown in Fig.~\ref{fig:teaser}.
    \item Extensive experiments demonstrate that our method achieves SOTA performance not only on benchmark datasets but also on challenging in-the-wild images captured by mobile devices, exhibiting superior generalization across diverse real-world reflection scenarios. 
\end{itemize}

\section{Related Work}
\label{sec:related}

\subsection{Reflection Removal}
Reflection removal is an ill-posed problem that aims to separate the reflection layer from a mixed image and recover the underlying transmission layer. It is well-established that additional information can significantly simplify this task. For instance, multi-view~\cite{niklaus2021learned, li2020improved, guo2014robust} information can guide robust reflection removal, but capturing image sequences is often redundant. Flash-based methods~\cite{lei2023robust, wang2024flash} utilize flash/no-flash image pairs, while polarization-based techniques~\cite{lei2020polarized} exploit different polarization angles. However, these approaches rely on specialized equipment, limiting their general applicability. As a result, we focus on the most challenging yet widely applicable task: single-image reflection removal (SIRR).

Traditional methods for reflection removal rely on handcrafted priors, such as the assumption that reflections are blurred~\cite{li2014single, wan2016depth} or exhibit ghosting effects~\cite{shih2015reflection}. However, these assumptions frequently break down in complex real-world scenarios, resulting in suboptimal performance. Universal image restoration techniques~\cite{chen2022simple, kawar2022denoising, wang2022uformer} could serve as a potential solution, however, they cannot outperform methods designed specifically for single image reflection removal. In recent years, learning-based methods~\cite{zhao2025reversible, hong2024differ, hu2023single, hu2024single, song2023robust, zhong2024language, dong2021location, hu2021trash, fan2017generic, wei2019single, yang2018seeing, wang2024promptrr} have become the mainstream approach, enabling models to remove reflections by training on large datasets. While some real datasets~\cite{wan2017benchmarking, zhang2018single, li2020single, zhu2023Weather} have been collected, their scale and quality remain insufficient. To address this, some methods employ empirical formulations~\cite{hu2023single}, physics-based rendering~\cite{kim2020single}, and tailored models~\cite{wen2019single}to generate synthetic data. Nevertheless, these approaches often struggle to bridge the gap between synthetic and real-world data. In this work, we introduce a comprehensive data collection pipeline to acquire a large volume of real-world data DRR and a data filtering strategy to enhance the quality of synthetic data.

% \begin{table*}[t]
% \centering
% % \resizebox{\columnwidth}{!}{%
% \caption{
% Summary of existing real reflection datasets.
% }
% \fontsize{8}{10}\selectfont
% \begin{tabular}{l||ccccccc}
% \toprule
% Dataset & Year & Videos & One-to-many angle & Usage & Pairs number & Scenes & Average resolution \\
% \midrule
% $SIR^2$\cite{wan2017benchmarking} & 2017 & $\times$ & $\times$ & Test & 500 & 126 & 540*400 \\
% $Real$\cite{zhang2018single} & 2018 & $\times$ & $\times$ & Train/Test& 89/20 & 89/20 & 1152*930 \\
% $Nature$\cite{li2020single} & 2020 & $\times$ & $\times$ & Train/Test & 200/20 & 68 & 598*398 \\
% $RRW$\cite{zhu2023Weather} & 2023 & \checkmark & $\times$ & Train & 14952 & 150 & 2580*1460 \\
% $DRR(Ours)$ & 2025 & \checkmark & \checkmark & Train/Test & \textbf{23303}/\textbf{400} & \textbf{217}/\textbf{40} & \textbf{3840*2160} \\
% \bottomrule
% \end{tabular}

% % }

% \vspace{-0.05in}
% \label{tab:dataset_comparison}
% \end{table*}

% diversity:
% 1: scene diversity(number, class)
% 2: reflection diversity
% (number, angle, content, intensity)
% 3. combination(one-to-many)

\begin{table}[t]
\centering
\caption{Summary of existing real reflection datasets. Compared to these datasets, our proposed DRR dataset demonstrates significant advantages in three key aspects: (1) greater diversity in reflection (varying glass angles, reflected contents, reflection intensity) and scenes (various environment conditions), (2) a substantially larger collection of image pairs, and (3) superior image quality with higher resolution.}
\fontsize{8}{10}\selectfont
\setlength{\tabcolsep}{2pt}
\begin{tabular}{l|ccccc}
\toprule
 % & \textbf{SIR\textsuperscript{2}\cite{wan2017benchmarking}} & \textbf{Real\cite{zhang2018single}} & \textbf{Nature\cite{li2020single}} & \textbf{RRW\cite{zhu2023Weather}} & \textbf{DRR (Ours)} \\
  & \textbf{SIR\textsuperscript{2}} & \textbf{Real} & \textbf{Nature} & \textbf{RRW} & \textbf{DRR (Ours)} \\
\midrule
Year & 2017 & 2018 & 2020 & 2023 & 2025 \\
Videos & $\times$ & $\times$ & $\times$ & \checkmark & \checkmark \\
Angles & $\times$ & $\times$ & $\times$ & $\times$ & \checkmark \\
Usage & Test & Train/Test & Train/Test & Train & Train/Test \\
Pairs & 500 & 89/20 & 200/20 & 14952 & \textbf{23303}/\textbf{400} \\
Scenes & 126 & 89/20 & 68 & 150 & \textbf{217}/\textbf{40} \\
Avg. res. & 540*400 & 1152*930 & 598*398 & 2580*1460 & \textbf{3840*2160} \\
\bottomrule
\end{tabular}
\vspace{-1.5em}
\label{tab:dataset_comparison}
\end{table}

\subsection{Diffusion Model}
Diffusion models~\cite{ho2020denoising, rombach2022high, song2020denoising} have emerged as a powerful framework for image generation and restoration, leveraging an iterative denoising process that learns the underlying data distribution. Their ability to produce high-quality, diverse outputs has garnered significant attention in recent years. Notably, diffusion models have been successfully applied to a wide range of image restoration tasks~\cite{kawar2022denoising}, such as deblurring~\cite{whang2022deblurring}, super-resolution~\cite{wang2024sinsr, wang2024exploiting},  inpainting~\cite{lugmayr2022repaint, xie2023smartbrush} and LLIE~\cite{zhou2025low, lan2025efficient, yi2023diff}, demonstrating exceptional performance in recovering fine-grained details. These characteristics make diffusion models particularly well-suited for the challenging task of reflection removal, where disentangling complex visual structures and maintaining image fidelity are critical. Recent work, such as L-DiffER~\cite{hong2024differ}, utilizes ControlNet~\cite{zhang2023adding} to inject information from the mixed image and iteratively denoise it. However, this approach requires multiple steps to recover the transmission layer and text prompt to guide the process. To address this limitation, we propose an alternative solution based on a recently proposed one-step denoising strategy~\cite{ye2024stablenormal, xu2024matters}. It enables stable and deterministic results while significantly accelerating the process, making it more practical for real-world applications.

\section{Method}
\label{sec:method}

% In this section, we first present our data collection pipeline, which captures diverse real-world data and generates high-quality synthetic data. Next, we introduce our diffusion-based framework specifically designed for reflection removal. Finally, we propose a reflection-invariant finetuning strategy to fully leverage the diversity of our dataset and enhance robustness for generalizing to real-world reflection scenarios.

\subsection{Data Collection Pipeline}
\begin{figure}[t]
\captionsetup{type=figure}
\centering
\resizebox{\linewidth}{!}{
\includegraphics[width=1.0\textwidth]{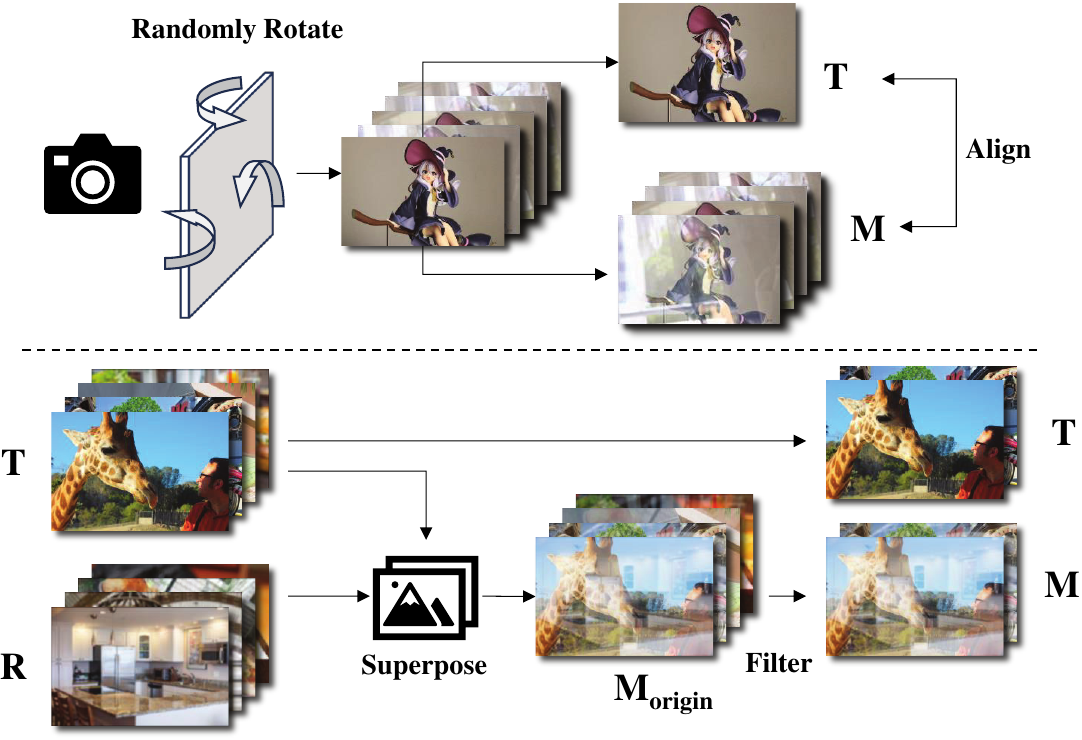}}
\caption {Data collection pipeline of real (above) and synthetic (below) data. Real data is captured by recording videos while rotating a glass panel at various angles, then processed to align mixed images with their ground truth transmission layers. Synthetic data is generated by randomly chosen coefficients and filtered to produce high-quality image pairs.}
\label{fig: collection}
\end{figure}

%define R M T
Formally, we define the target scene as transmission layer \textbf{T}, which is superposed with undesired reflection layer \textbf{R}, resulting in the mixed image \textbf{M}. Given that the core objective is to  translate mixed image \textbf{M} to transmission layer \textbf{T}, we design a comprehensive data collection pipeline to acquire aligned image pairs of \textbf{M} and \textbf{T}, including our diverse real-world data, and synthetic data serving as a supplementary, demonstrated in Fig.~\ref{fig: collection}.

\paragraph{Real Data}

Our dataset is captured using a Nikon Z50 camera and three mobile phones mounted on a fixed tripod, with a portable glass slab positioned in front of the lens. By rotating the glass at different angles within a sequence, we generate reflection images with varying reflection and intensity. The corresponding ground-truth transmission images are acquired by removing the glass entirely. To ensure diversity and robustness, we carefully vary key parameters, including scenes (indoor, outdoor, object-centric), lighting conditions (skylight, sunlight, incandescent), glass thickness (3 mm and 8 mm), camera-to-glass distance, viewing angles, exposure values, and aperture settings.

The dataset consists of 257 unique scenes, each captured with two glass thicknesses to ensure reflection diversity. It is partitioned into a training set (217 scenes, 23,303 image pairs) and a testing set (40 scenes, 400 image pairs). All training images are captured in 4K resolution (3840 × 2160 pixels) to provide high visual fidelity for model training. The testing set is further divided into two subsets: DRR-S, containing standard reflections captured with the same camera used in training, and DRR-C, containing challenging reflections captured using three mobile phones. This division enables a comprehensive evaluation of model performance across varying reflection complexities and real-world scenarios.

To address spatial shifts caused by glass refraction, we employ a robust post-processing pipeline. Inspired by~\cite{wan2017benchmarking}, we use Scale-Invariant Feature Transform (SIFT)~\cite{lowe2004distinctive} for feature point detection and Random Sample Consensus (RANSAC) for precise alignment between reflection images and their ground-truth transmission pairs. 

Compared to existing datasets, our collection offers superior image quality, substantial quantity, and exceptional diversity, derived from varying glass angles, reflected contents, reflection intensity and various environment conditions. Our method has faster capturing speed and is easier to scale up. The existing RRW dataset focuses on object movement and occlusions in front of the glass, lacking the variation in angular perspectives and diverse reflection content from the real scenes that characterize our dataset. A detailed comparison is provided in Table~\ref{tab:dataset_comparison}.

\paragraph{Synthetic Data Supplementary}
% In addition to datasets collected from the real world, synthetic data can serve as a significant supplementary resource to enrich the diversity of training data. In our work, we adopt the formulation from DSRNet~\cite{hu2023single} to generate synthetic data.
% % \begin{equation}
% % \textbf{M} = \gamma_1 \textbf{T} + \gamma_2 \textbf{R} - \gamma_1\gamma_2 \textbf{T} \circ \textbf{R},
% % \end{equation}
% % 
% We randomly sample transmission and reflection layers from the COCO~\cite{lin2014microsoft} and PASCAL VOC~\cite{everingham2015pascal} datasets.
% % and assign random values to the coefficients $\gamma_1 \in [0.8, 1.0]$ and $\gamma_2 \in [0.4, 1.0]$ during synthesis.

% An intuitive observation is that the synthetic data exhibit a wide range of quality: while some images closely resemble real-world reflections, others appear less realistic. To address this, we leverage CLIP similarity \cite{radford2021learning} with the text prompt ``image with glass reflection" to evaluate the synthetic data. Based on this metric, we filtered synthetic data from initially 69,443 pairs to 20,833 high quality pairs. Our synthetic data is also capable of generating multiple reflection for a single scene, thereby enhancing the diversity and practicality of the training data.

To complement real-world datasets, we generate synthetic data to enhance training diversity. Following DSRNet~\cite{hu2023single}, we synthesize mixed images from transmission and reflection layers, then use CLIP~\cite{radford2021learning} to filter for realism, retaining 20,833 high-quality pairs. Our synthetic data also supports multiple reflections per scene, further enriching the training set.

\subsection{Framework}

\begin{figure}[t]
\captionsetup{type=figure}
\centering
\resizebox{\linewidth}{!}{
\includegraphics[width=1.0\textwidth]{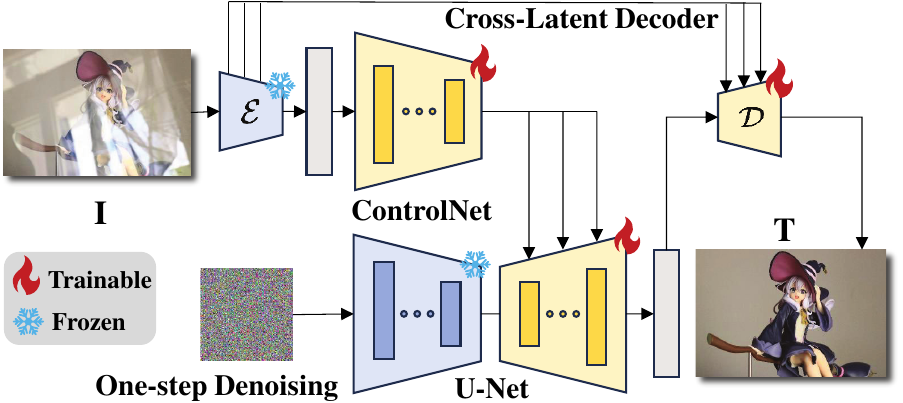}}
\caption {Our proposed framework. It consists of a U-net with one-step denoising strategy, a ControlNet to input the mixed image processed by the encoder $\mathcal{E}$, and a cross-latent decoder $\mathcal{D}$ to mitigate blurriness and preserve details.}
\label{fig: framework}
\end{figure}

Our proposed dataset offers a comprehensive and diverse training resource, however, it is impractical to encompass all possible reflection types. Fortunately, diffusion models provide an effective solution. By leveraging the powerful generative priors of diffusion models, we can maximize the utility of limited data and address more challenging reflection scenarios. Our diffusion-based model is composed of three key parts, a U-net for one-step denoising, a ControlNet to input the mixed image, and a cross-latent decoder to preserve details, Shown in Fig.~\ref{fig: framework}.

Traditional diffusion models generate results through iterative denoising from Gaussian noise. But we use a one-step denoising strategy as an alternative, the target latent is predicted with only one step by the U-net, resulting in deterministic results and fast inference. ControlNet \cite{zhang2023adding} is initially designed to impose structural constraints on generated images. In our work, we adapt it as a mechanism to inject information from mixed images. To address the inherent limitations of one-step diffusion, such as over-smoothed results and potential shifts in details during the denoising process, we add a cross-latent decoder $\mathcal{D}$ to preserve the high-frequency details of the input mixed image. Inspired by~\cite{wang2024flash, ronneberger2015u}, the multi-scale latent features extracted by the encoder are directly connected to the decoder with zero convolution, creating a shortcut that preserve the initial information of the mixed image.

\subsection{Progressive Training}
\begin{figure}[t]
\captionsetup{type=figure}
\centering
\includegraphics[width=\columnwidth]{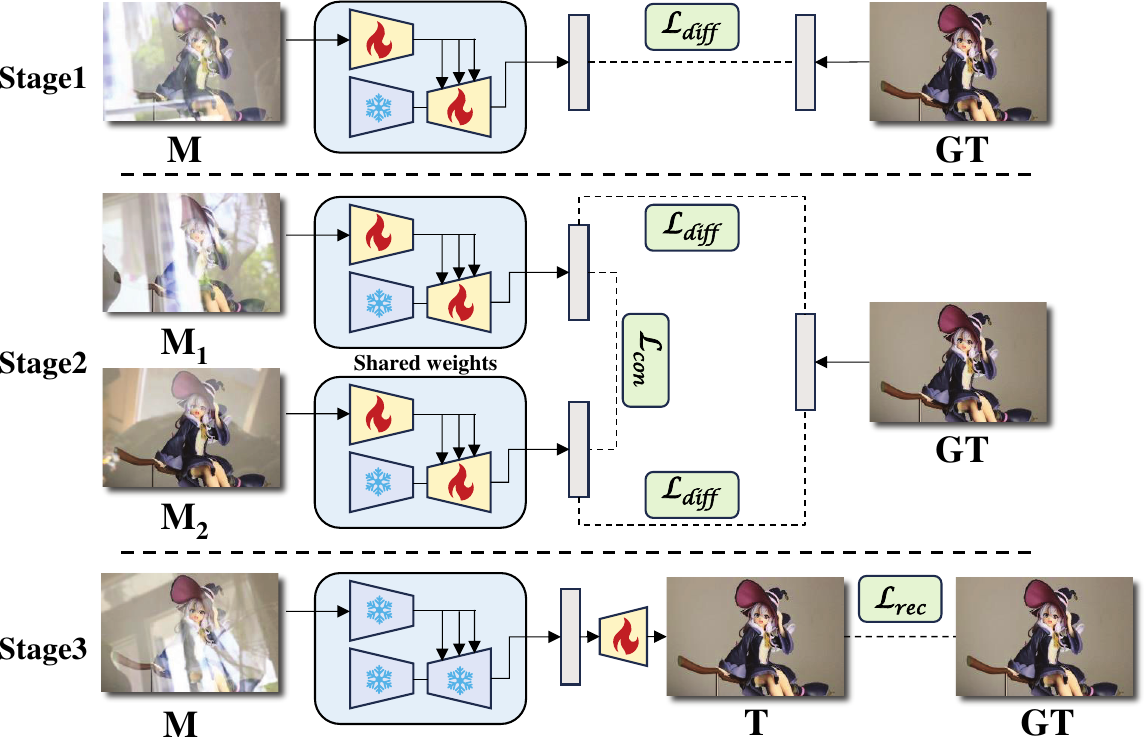} % Half the column width
\caption{The three stages of progressive training. First, we train the ControlNet and the upsampling blocks of the U-Net using the basic one-step diffusion loss. Second, we finetune these components by incorporating the consistent loss. Finally, we train the cross-latent decoder using the image reconstruction loss.}
\label{fig:training_process}
\end{figure}

To effectively train our reflection removal model, we adopt a progressive training strategy that decomposes the learning process into three distinct yet interconnected phases, as illustrated in Fig.~\ref{fig:training_process}. This hierarchical approach enables stable optimization and ensures that each component of our model is properly trained to handle the complex task of reflection removal.

\noindent\textbf{Foundation Training}
The initial stage focuses on establishing the fundamental reflection removal capability by jointly training the ControlNet and the upsampling blocks of the U-Net. We employ the following one-step diffusion loss function:
\begin{equation}
\mathcal{L}_{diff} = \mathbb{E}_{z_t, \textbf{M}} \Big[ \|z_t - \mu_{\theta, \phi}^{z_t}(\textbf{M})\|_2^2 \Big],
\label{eq: loss_diff_}
\end{equation}
where $\theta$ and $\phi$ refer to the parameters of U-Net and ControlNet, $z_t$ is the latent representation after adding noise over $t$ steps with $t \in (0, T)$. (in Stable Diffusion, T is set to 1,000). For notational simplicity, we leave the full formulation in the supplementary material.

\noindent\textbf{Reflection-Invariant Fine-tuning}
Building upon the foundation, we introduce a novel reflection-invariant fine-tuning strategy that leverages the unique characteristics of our dataset, where each scene contains multiple mixed images with varying reflections. Our key insight is that the model should produce consistent results regardless of the specific reflection patterns present in the input. This is achieved by incorporating a consistent loss:
\begin{equation}
\mathcal{L}_{con} = \mathbb{E}_{\textbf{M}_1, \textbf{M}_2} \Big[ \|\mu_{\theta,  \phi}^{z_t}(\textbf{M}_1) - \mu_{\theta, \phi}^{z_t}(\textbf{M}_2)\|_2^2 \Big],
\end{equation}
\begin{equation}
    \mathcal{L} = \mathcal{L}_{diff}(\textbf{M}_1) + \mathcal{L}_{diff}(\textbf{M}_2) + \mathcal{L}_{con}(\textbf{M}_1, \textbf{M}_2).
\end{equation}

The reflection-invariant finetuning enhances generalization by focusing on invariant properties of transmission scenes rather than variable reflection characteristics, making the model more robust and generalize to complex reflection scenarios.

\noindent\textbf{Cross-Latent Decoder}
% In the final stage, we freeze the previously trained components and focus on training the cross-latent decoder to mitigate blurriness and preserve high-frequency details. This is accomplished through a comprehensive image reconstruction loss:
% \begin{equation}
% \mathcal{L}_{rec} = \mathcal{L}_1 + \mathcal{L}_{SSIM} + \mathcal{L}_{LPIPS}.
% \end{equation}
In the final stage, we freeze the previously trained components and focus on training the cross-latent decoder to mitigate blurriness and preserve high-frequency details. This is accomplished through a comprehensive image reconstruction loss:
\begin{equation}
\mathcal{L}{rec} = \mathcal{L}1 + \lambda \cdot (\mathcal{L}_{SSIM} + \mathcal{L}_{LPIPS}),
\end{equation}
where $\lambda = 0.2$ is the weighting coefficient for structural and perceptual loss terms.

\begin{figure*}[t]
\captionsetup{type=figure}
\centering
\resizebox{\linewidth}{!}{
\includegraphics[width=1.0\textwidth]{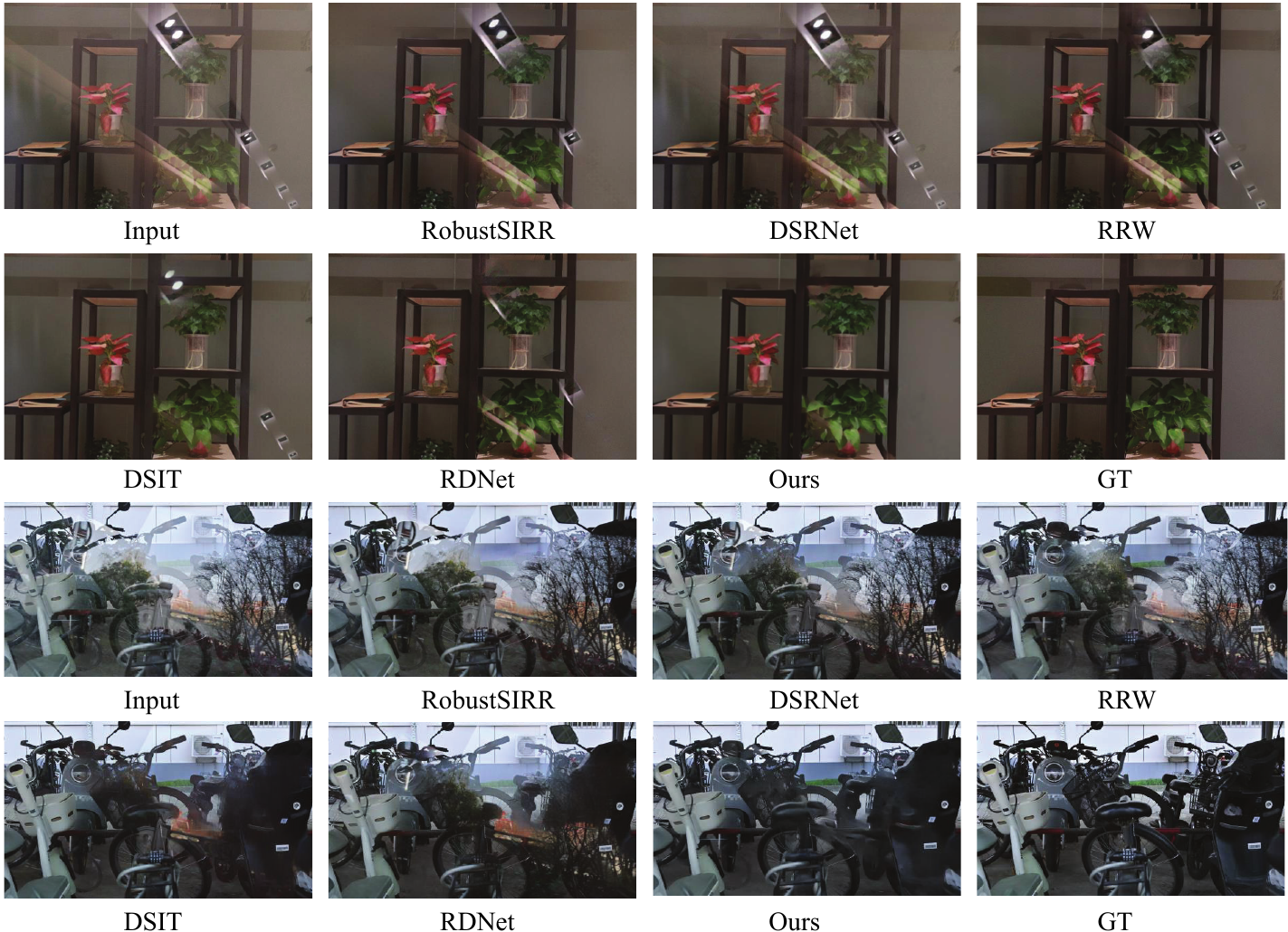}}
\caption {Qualitative comparison of our method with other approaches, demonstrating superior performance in preserving background details while effectively removing reflections.}
\label{fig: qualitative_comparison}
\end{figure*}

\begin{table*}[t]
\centering
\caption{Quantitative comparisons on the existing three reflection benchmarks, and our new dataset DRR, consisting a standard set (DRR-S) and a challenging set (DRR-C). The scores after finetuning on our data setting are labeled as $\dagger$. The best results are in \textbf{bold}, and the second-best results are \underline{underlined}.}
\label{tab:metrics}
\fontsize{9}{10}\selectfont
\begin{tabular}{lr|cccccc|cccc}
\toprule
\multirow{2}{*}{Methods} & \multirow{2}{*}{Venue} & \multicolumn{2}{c}{$Nature$ (20)} & \multicolumn{2}{c}{$Real$ (20)} & \multicolumn{2}{c|}{$SIR^{2}$ (500)} & \multicolumn{2}{c}{$DRR$-$S$ (200)} & \multicolumn{2}{c}{$DRR$-$C$ (200)} \\ \cmidrule(l{2pt}r{2pt}){3-4} \cmidrule(l{2pt}r{2pt}){5-6} \cmidrule(l{2pt}r{2pt}){7-8} \cmidrule(l{2pt}r{2pt}){9-10} \cmidrule(l{2pt}r{2pt}){11-12}
 & & PSNR$\uparrow$ & SSIM$\uparrow$ & PSNR$\uparrow$ & SSIM$\uparrow$ & PSNR$\uparrow$ & SSIM$\uparrow$ & PSNR$\uparrow$ & SSIM$\uparrow$ & PSNR$\uparrow$ & SSIM$\uparrow$ \\ \midrule
RobustSIRR & CVPR 2023 & 20.94 & 0.770 & 22.71 & 0.787 & 22.61 & 0.872&  19.68         &    0.756   & 20.24 & 0.692  \\
DSRNet & ICCV 2023 & 24.86 & 0.823 & 23.31 & 0.791 & 25.65 & 0.919 &  22.33         &    0.846   & 21.93 & 0.820 \\
RRW & CVPR 2024 & 25.79 & 0.833 & 21.51 & 0.767 & 25.31 & 0.907 &  22.39         &    0.857   & 21.84 & 0.820 \\
L-DiffER & ECCV 2024 & 23.95 & 0.831 & 23.77 & 0.821 & 25.18 & 0.911 & - & - & - & - \\
DSIT & NeurIPS 2024 & \underline{26.25} & 0.833 & 24.54 & 0.814 & 26.34 & 0.922 &  \underline{23.48}         &    \underline{0.869}   & \underline{22.46} & 0.817  \\
RDNet & CVPR 2025 & 25.94 & \underline{0.835} & \underline{25.10} & \underline{0.833} & \underline{26.38} & \underline{0.924} &  23.31         &    0.865   & 22.25 & \underline{0.823}  \\ \midrule
Ours & - & \textbf{27.05} & \textbf{0.846} & \textbf{25.24} & \textbf{0.840} & \textbf{27.32} & \textbf{0.931} &  \textbf{27.21}         &    \textbf{0.901}   & \textbf{23.83} & \textbf{0.840}  \\ \bottomrule
\end{tabular}
\end{table*}
\begin{table*}[t]
\centering
\caption{Quantitative results of ablation study, CLD refers to Cross-Latent Decoder, DRR is our dataset Diverse Reflection Removal, RIF refers to Reflection-Invariant Finetuning.}
\label{tab:ablation_metrics}
% \fontsize{8}{10}\selectfont
\begin{tabular}{lll|cccccc|cccc}
\toprule
\multirow{2}{*}{CLD} & \multirow{2}{*}{DRR} & \multirow{2}{*}{RIF} & \multicolumn{2}{c}{$Nature$  (20)} & \multicolumn{2}{c}{$Real$ (20)} & \multicolumn{2}{c}{$SIR^{2}$ (500)} & \multicolumn{2}{|c}{$DRR$-$S$ (200)} & \multicolumn{2}{c}{$DRR$-$C$ (200)}\\  
\cmidrule(l{2pt}r{2pt}){4-5} \cmidrule(l{2pt}r{2pt}){6-7} \cmidrule(l{2pt}r{2pt}){8-9} \cmidrule(l{2pt}r{2pt}){10-11} \cmidrule(l{2pt}r{2pt}){12-13}
                         &&& PSNR$\uparrow$ & SSIM$\uparrow$ & PSNR$\uparrow$ & SSIM$\uparrow$ & PSNR$\uparrow$ & SSIM$\uparrow$ & PSNR$\uparrow$ & SSIM$\uparrow$ & PSNR$\uparrow$ & SSIM$\uparrow$ \\ \midrule
$\times$ & $\times$ & $\times$                     & 25.54& 0.782 & 23.67 & 0.762 & 24.96 & 0.861 & 22.32            & 0.769    & 20.50 & 0.666       \\ 
\checkmark & $\times$ & $\times$                 &     26.70      &    0.834       &    23.99       &   0.828        &     25.98      &  0.921         &       24.43        &   0.881   & 22.51 &  0.828     \\ 
\checkmark & \checkmark &  $\times$                 & 26.71 & 0.839 & 24.81 & 0.836 & 26.78 & 0.927 &    26.44            &     0.898   & 23.34 &  0.835      \\ 
\checkmark & \checkmark & \checkmark                   & \textbf{27.05} & \textbf{0.846} & \textbf{25.24} & \textbf{0.840} & \textbf{27.32} & \textbf{0.931} &     \textbf{27.21}           &     \textbf{0.901}    & \textbf{23.83} & \textbf{0.840}       \\ \bottomrule
\end{tabular}
\end{table*}

% - 
% - C
% - D C
% - D F C

\begin{figure}[H]
\captionsetup{type=figure}
\centering
\includegraphics[width=1.0\columnwidth]{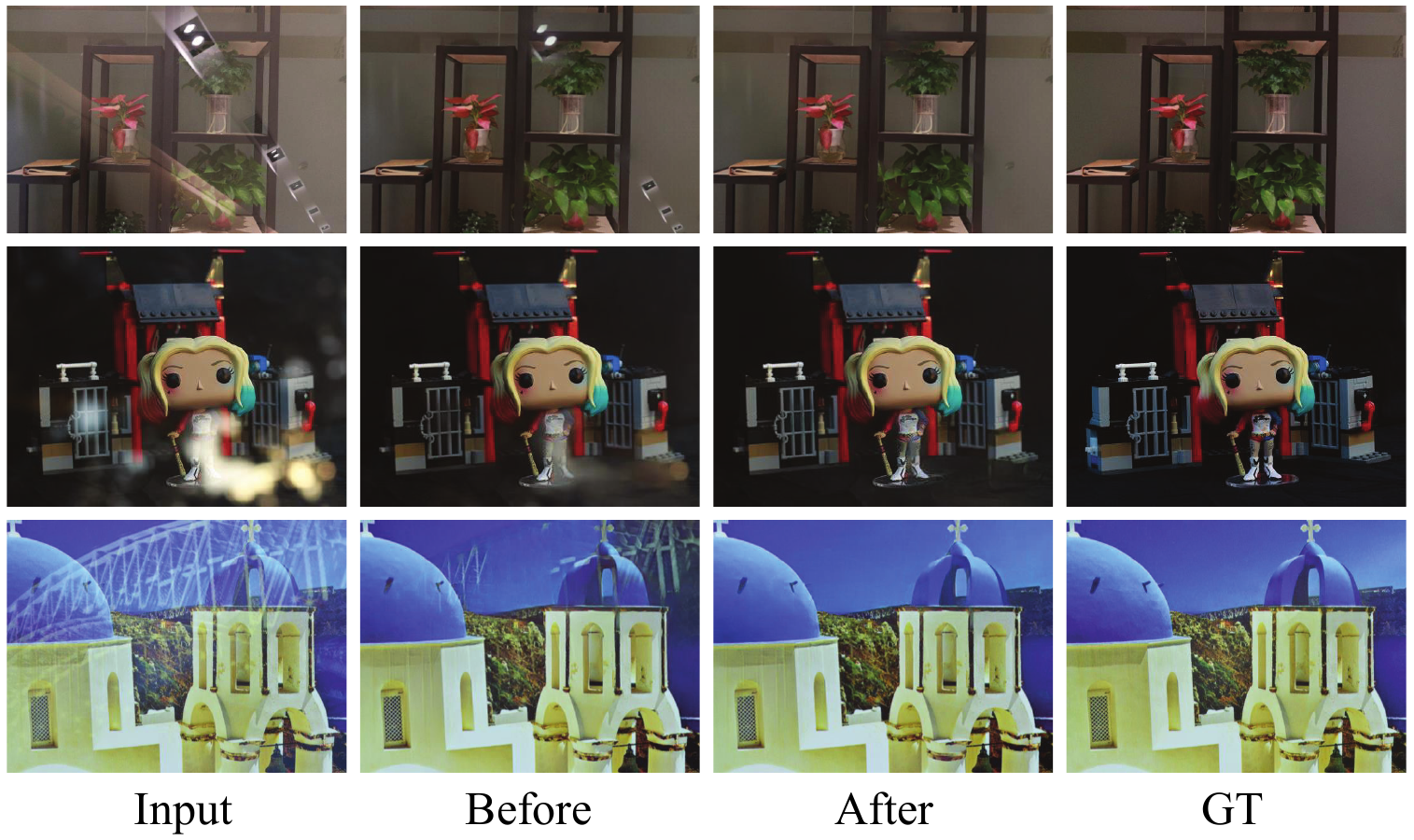}
\caption{Visual comparison of DSIT performance before and after fine-tuning on our dataset. The results demonstrate significant improvements in reflection removal quality after adaptation to our
data setting.}
\label{fig:finetune_dsit}
\end{figure}

\section{Experiments}
\label{sec:experiments}

\subsection{Implementation Details}

Our model is trained on a combination of datasets, including our new DRR dataset, synthetic data, and established benchmarks (\textit{Real} and \textit{Nature}). We employ standard data augmentation techniques and a progressive 3-stage training strategy (4 days total on 3$\times$NVIDIA GeForce RTX 3090 GPUs) initialized with Stable Diffusion v2.1 weights. Evaluation covers three standard benchmarks plus our new DRR benchmark with 400 image pairs across 20 scenes. Our method achieves fast inference speed, processing a 768-resolution image in approximately  1 second on a single RTX 3090 GPU.

\subsection{Comparison to State-of-the-arts}
We compare our method with several recently proposed approaches, including RobustSIRR~\cite{song2023robust}, DSRNet~\cite{hu2023single}, L-DiffER~\cite{hong2024differ}, RRW~\cite{zhu2023Weather}, DSIT~\cite{hu2024single} and RDNet~\cite{zhao2025reversible}. All evaluations are conducted using the same testing scripts and the same testing data to ensure fairness. As L-DiffER's pre-trained models and code are not available, we use the results from the original paper. Qualitative results in Fig.~\ref{fig: qualitative_comparison} show that our diffusion-based framework show SOTA performance.
% old benchmark
The quantitative results on the three established benchmarks are demonstrated in Table~\ref{tab:metrics}. Our method achieves the highest average PSNR and SSIM scores, surpassing all recent approaches. This demonstrates the overall superiority of our approach. 

% new benchmark
In addition, we evaluate the performance on our two testing benchmarks. The quantitative results are also presented in Table~\ref{tab:metrics}. We observe that existing methods perform poorly on these benchmarks, suggesting potential overfitting to older datasets. In contrast, our approach exhibits superior generalization capabilities, achieving strong performance on both the standard and challenging sets.

% finetuning comparison
% Recognizing that all compared methods exhibit improved performance after finetuning on our dataset
To explore the superiority of our dataset, we fine-tune all the methods in our data setting, and present a comprehensive comparison of their results before and after this adaptation in Fig.~\ref{fig:finetune_dsit}. It indicates that our dataset significantly enhances the robustness of reflection removal models. To illustrate this enhancement, we specifically select DSIT as an example.

\subsection{Ablation Study}
\begin{figure*}[!htbp]
\captionsetup{type=figure}
\centering
\resizebox{\linewidth}{!}{
\includegraphics[width=1.0\textwidth]{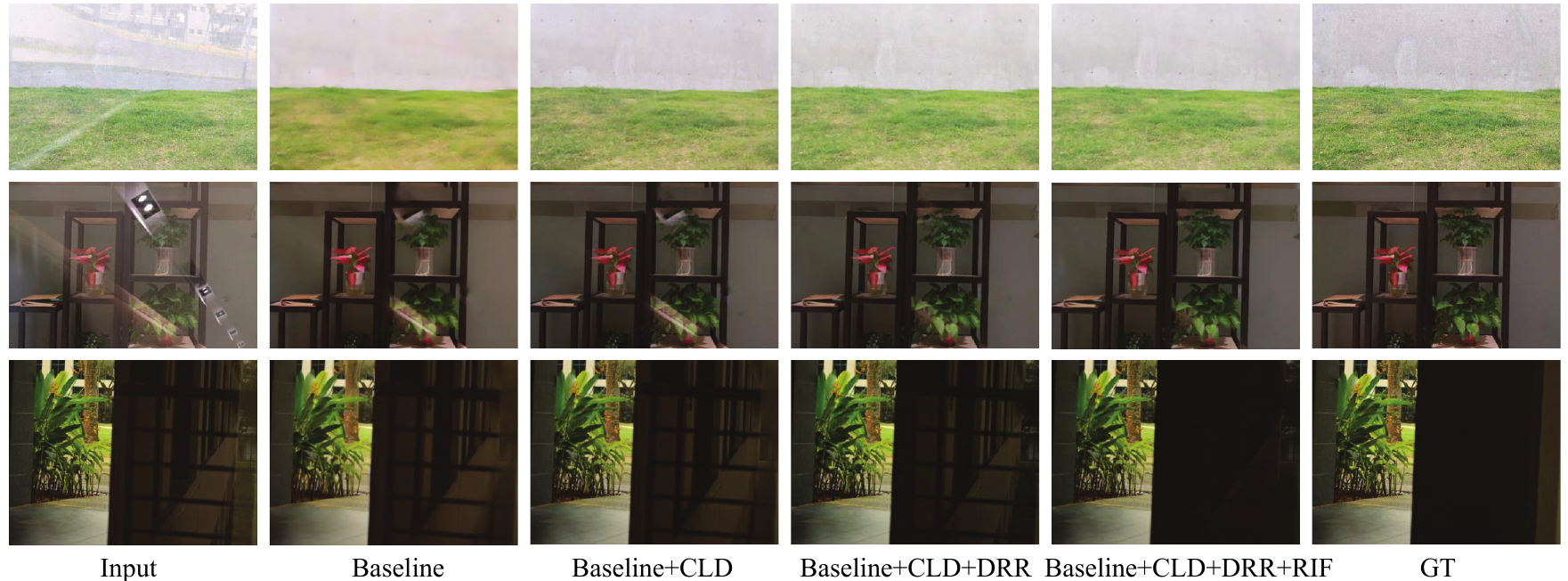}}
\caption {Qualitative results of the ablation study. CLD is Cross-Latent Decoder, DRR is our dataset Diverse Reflection Removal, RIF refers to Reflection-Invariant Finetuning. Baseline is the basic setting without CLD, DRR and RIF.}
\label{fig: ablation}
\end{figure*}

To validate the effectiveness of each component in our proposed method, we conduct a comprehensive ablation study. We evaluate the impact of 1) the cross-latent decoder module, 2) our new dataset DRR, and 3) the reflection-invariant finetuning strategy by adding the component gradually from a baseline. The results are summarized in Table~\ref{tab:ablation_metrics} and visual
quality is demonstrated in Fig.~\ref{fig: ablation}

\paragraph{Cross-Latent Decoder}
The baseline employs a vanilla decoder from Stable Diffusion without connecting it with encoder, thus failing to restore high-frequency details. To address the over-smooth issue of one-step diffusion process, we consider cross-latent decoder as an essential component. After incorporating this module, more original information is directly injected during the decoding process, effectively mitigating blurring effects and preserving details. This enhancement results in more faithful reconstruction quality, with improved PSNR and SSIM scores.

% \begin{figure}[H]
% \captionsetup{type=figure}
% \centering
% \resizebox{\linewidth}{!}{
% \includegraphics[width=1.0\textwidth]{figs_cmyk/wild.pdf}}
% \caption {Our method generalizes robustly to diverse real-world reflections, including reflections from water surfaces, glossy plastics, digital displays, and even stylized scenes such as anime.}
% \label{fig:de_any_image}
% \end{figure}

\paragraph{Influence of Dataset}
When training without our newly proposed dataset DRR, the model relies only on the old real datasets (Nature and Real) and synthetic data. Our experiments reveal that incorporating DRR enhances the performance across all benchmarks, including both established benchmarks and our proposed ones. This demonstrates that DRR introduces diverse and challenging scenarios, which are crucial for improving the generalization and robustness of reflection removal models.

\paragraph{Reflection-Invariant Finetuning}
Initially, we train the model on the entire dataset by treating all image pairs equally, without any distinction. However, experiments confirm that it is not a ideal solvement, thus leading to suboptimal performance. After incorporating our reflection-invariant finetuning strategy, our model become more robust in handling diverse reflection scenarios, ultimately achieving more stable and reliable results.

% \subsection{Dereflection Any Image}
% To further demonstrate the versatility and generalization capability of our proposed method, we conduct additional experiments on challenging real-world images exhibiting various complex reflection patterns, such as reflections on water surfaces, glossy plastics, digital screens, and even stylized anime scenes. Due to the absence of ground-truth transmission layers in these scenarios, our evaluation is qualitative. Visualization results in Fig.~\ref{fig:de_any_image} clearly illustrate the robustness and effectiveness of our approach in handling diverse, uncontrolled reflection conditions.

\section{Conclusion}
\label{sec:conclusion}

We present a comprehensive framework for single-image reflection removal, comprising a high-quality dataset (DRR) featuring diverse reflection scenarios and an efficient capture pipeline, a robust diffusion-based model tailored for reflection removal, and a novel reflection-invariant finetuning strategy to enhance generalization. Our dataset significantly improves training quality by addressing the limitations of existing datasets. Our method achieves state-of-the-art performance on both common and our newly introduced benchmarks. Extensive experiments and ablation studies validate the effectiveness of each component, demonstrating superior generalization capabilities and robustness across diverse real-world scenarios. This work not only advances reflection removal technology but also provides a valuable resource and a strong baseline for future research in the field.

\noindent\textbf{Limitation} 
% % As reflection removal is an ill-posed problem, it becomes particularly challenging when the reflection and target scene are seamlessly superposed. 
When the reflection and target scene are seamlessly integrated, it is hard to distinguish which one is the target to be preserved. Incorporating semantic or user guidance offers a promising direction for future research.

\bibliography{aaai2026}

% \section{Acknowledgment}
% This work was supported by the NSFC under Grant 62322604 and 62576207. The authors
% express gratitude to the anonymous reviewers for their valuable feedback. 

\appendix

\section{Details of Loss Functions}
% Latent diffusion models~\cite{ho2020denoising, rombach2022high}, 
In the forward diffusion process, Gaussian noise $\varepsilon$ is incrementally applied to the latent representation $z$ of the training image, resulting in a noisy latent representation defined as: $z_t = \sqrt{\bar{\alpha_t}} z + \sqrt{1 - \bar{\alpha_t}} \varepsilon$. In the backward denoising process, the traditional diffusion models learn to predict the noise at randomly sampled step $t$, and denoise with multi-step for inference. The corresponding multi-step loss is formulated as:
\begin{equation}
\mathcal{L}_{multi-step} = \mathbb{E}_{\epsilon \sim \mathcal{N}(0,1), c, t} \Big[ \|\epsilon - \mu_{\theta}^{\epsilon}(z_t, c, t)\|_2^2 \Big],
\end{equation}
where $z_t$ is the latent representation after adding noise over $t$ steps with $t \in (0, T)$. In Stable Diffusion, $T$ is set to 1,000. The embedding condition $c$ is derived from the text prompt ``remove glass reflection". $\mu$ is the U-net with parameters $\theta$, predicting the noise $\epsilon$ as the training target.

The noise introduced during the multi-step diffusion process creates challenges in producing stable results, making traditional diffusion models unsuitable for our reflection removal task, which requires deterministic outputs. In contrast, our approach employs one-step denoising \cite{ye2024stablenormal, xu2024matters}, which significantly accelerates inference speed while producing deterministic results. The denoising loss function for our method is formulated as:
\begin{equation}
\mathcal{L}_{one-step} = \mathbb{E}_{z_t, c, t} \Big[ \|z_t - \mu_{\theta}^{z_t}(z_T, c,  t)\|_2^2 \Big],
\end{equation}
 During the training process, the noise is added to the max step $T$, the target is predicting latents at $t$ step with $t \in (0, T)$. By setting $t = 0$ during inference, we obtain the final deterministic result.

We adapt ControlNet~\cite{zhang2023adding} as a mechanism to inject information from mixed images. By jointly training the ControlNet with the upsampling blocks of the U-Net architecture, we enable the model to effectively perform reflection removal. Our loss function is formulated as:
\begin{equation}
\mathcal{L}_{diff} = \mathbb{E}_{z_t, c, \textbf{M}, t} \Big[ \|z_t - \mu_{\theta}^{z_t}(z_T, c, t, f_\phi(\mathcal{E}(\textbf{M})))\|_2^2 \Big],
\end{equation}
where $\mu$ and $f$ denote U-Net and ControlNet with parameters $\theta$ and $\phi$, respectively. The mixed image \textbf{M} is first processed by the encoder $\mathcal{E}$, and the resulting encoded features are subsequently fed as input into the ControlNet. For simplicity, we express the formulation as follows:
\begin{equation}
\mathcal{L}_{diff} = \mathbb{E}_{z_t, \textbf{M}} \Big[ \|z_t - \mu_{\theta, \phi}^{z_t}(\textbf{M})\|_2^2 \Big],
\label{eq: loss_diff}
\end{equation}

We introduce a novel reflection-invariant fine-tuning strategy that encourages consistent outputs across varying reflection patterns. Specifically, we first randomly select two different mixed images, $\textbf{M}_1$ and $\textbf{M}_2$ from our real dataset DRR and synthetic data. then we use two models with shared weights to process them respectively. The outputs of these models are used to calculate the basic diffusion loss show in Eq.~\ref{eq: loss_diff}. Additionally, we design a novel consistent loss to constrain the outputs of the two models to be similar in the latent space, formulated as:
\begin{equation}
\mathcal{L}_{con} = \mathbb{E}_{\textbf{M}_1, \textbf{M}_2} \Big[ \|\mu_{\theta,  \phi}^{z_t}(\textbf{M}_1) - \mu_{\theta, \phi}^{z_t}(\textbf{M}_2)\|_2^2 \Big],
\end{equation}
We combine the consistent loss with the two basic diffusion losses using equal weights, formulated as:
\begin{equation}
    \mathcal{L} = \mathcal{L}_{diff}(\textbf{M}_1) + \mathcal{L}_{diff}(\textbf{M}_2) + \mathcal{L}_{con}(\textbf{M}_1, \textbf{M}_2),
\end{equation}

At the final stage of our progressive training, the cross-latent decoder is trained using a combination of L1 loss, SSIM \cite{wang2004image} loss, and LPIPS \cite{zhang2018unreasonable} loss, with equal weighting applied to each component:
\begin{equation}
\mathcal{L}_{rec} = \mathcal{L}_1 + \mathcal{L}_{SSIM} + \mathcal{L}_{LPIPS}.
\end{equation}

\section{More Details of DRR}
\begin{figure}[t]
\captionsetup{type=figure}
\resizebox{\linewidth}{!}{
\includegraphics[width=1.0\textwidth]{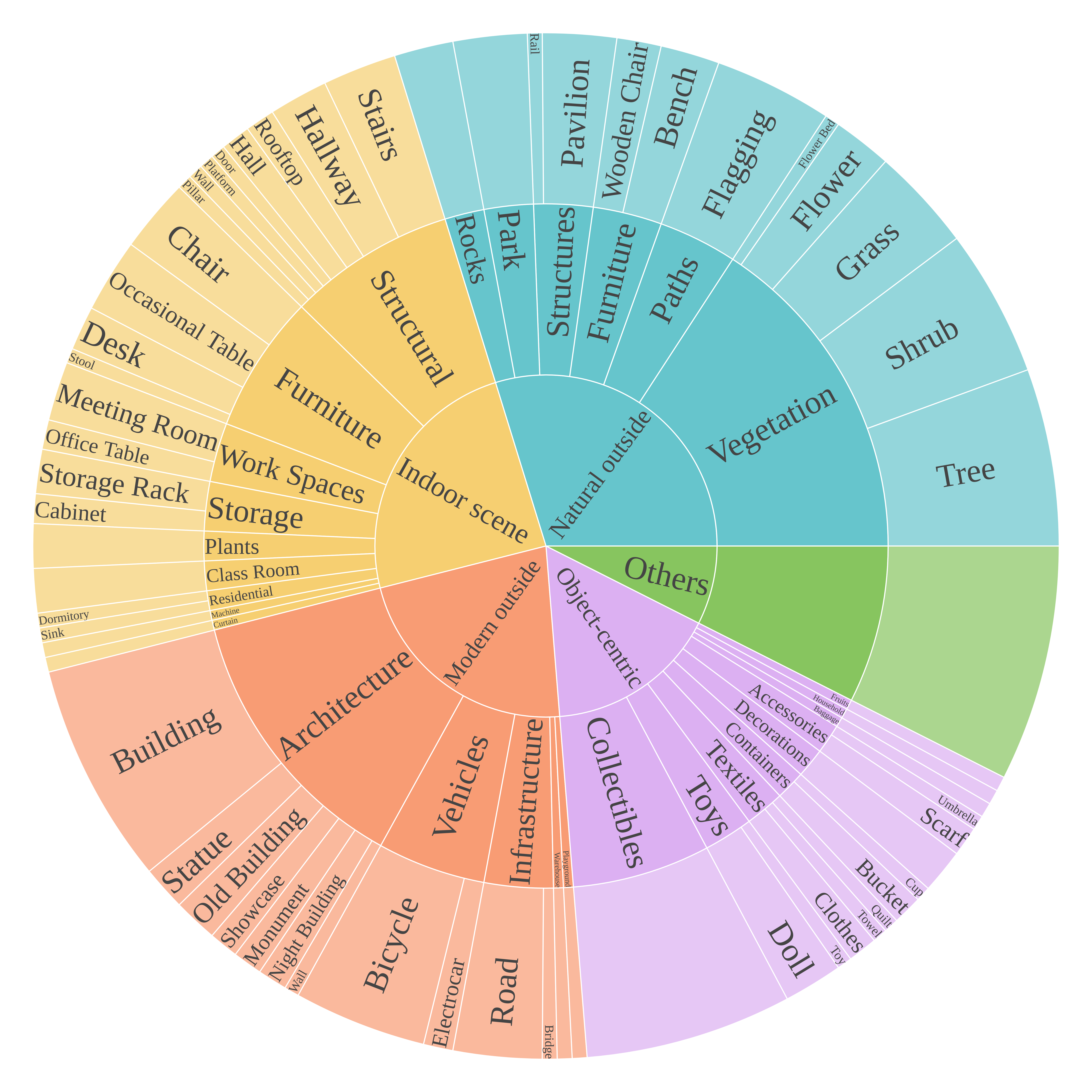}}
\caption {Scene types of DRR, highlighting
the balanced and comprehensive coverage of our data.}
\label{fig: pie}
\end{figure}

\begin{figure*}[t]
\captionsetup{type=figure}
\resizebox{\linewidth}{!}{
\includegraphics[width=1.0\textwidth]{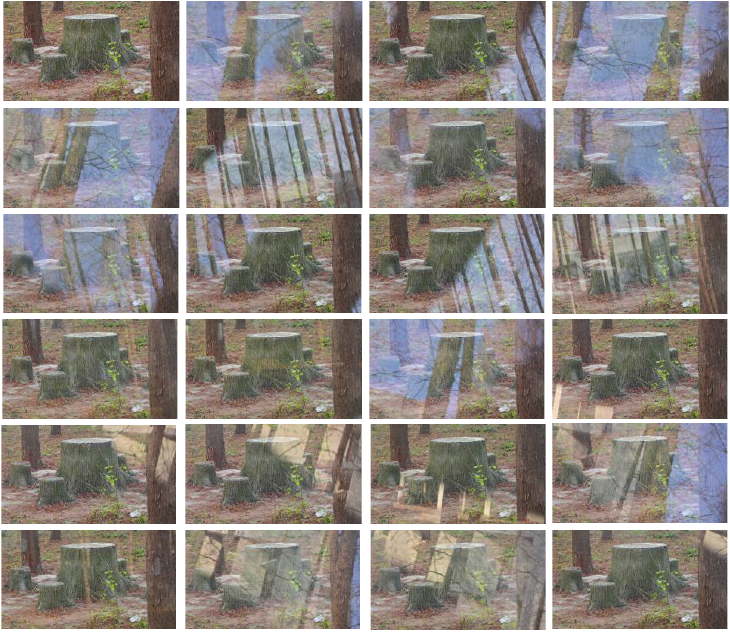}}
\caption {Additional visual demonstration of DRR. Each scene has diverse reflection patterns, varying glass angles and reflection intensity. The ground truth transmission layer is the first image while the remaining images represent various mixed images.}
\label{fig: visual}
\end{figure*}

To better illustrate the diversity of our dataset, we have included a pie chart (Fig.~\ref{fig: pie}) visualizing the distribution of scene types, ranging from indoor scenes, mordern and natural outdoor scenes and object-centric scenes. This highlights the balanced and comprehensive coverage of our data. To show the angle diversity of our dataset, we randomly select several mixed images of one scene, demonstrated in Fig.~\ref{fig: visual}. The ground truth transmission layer is the first image while the remaining images represent various mixed images.

\section{More Details of Synthetic Data}
In addition to datasets collected from the real world, synthetic data can serve as a significant supplementary resource to enrich the diversity of training data. In our work, we adopt the following formulation from DSRNet~\cite{hu2023single} to generate synthetic data:
\begin{equation}
\textbf{M} = \gamma_1 \textbf{T} + \gamma_2 \textbf{R} - \gamma_1\gamma_2 \textbf{T} \circ \textbf{R},
\end{equation}
where T, R, I represent the transmission, reflection and mixed layers, respectively. To enhance the diversity of the synthesized data, we randomly sample transmission and reflection layers from the COCO~\cite{lin2014microsoft} and PASCAL VOC~\cite{everingham2015pascal} datasets and assign random values to the coefficients $\gamma_1 \in [0.8, 1.0]$ and $\gamma_2 \in [0.4, 1.0]$ during synthesis.
An intuitive observation is that the synthetic data exhibit a wide range of quality: while some images closely resemble real-world reflections, others appear less realistic. To address this, we leverage CLIP similarity \cite{radford2021learning} with the text prompt ``image with glass reflection" to evaluate the synthetic data. Based on this metric, we filtered synthetic data from initially 69,443 pairs to 20,833 high quality pairs. Similar to our real-world dataset, our synthetic data is also capable of generating multiple reflection for a single scene, thereby enhancing the diversity and practicality of the training data.

\section{Implementation Details}

Our model is trained on a combination of datasets, including our newly introduced real dataset DRR, our synthetic dataset, and the previously established Real and Nature datasets. To enhance the robustness and generalization capability of our model, we implement a comprehensive suite of data augmentation techniques. These techniques include random cropping to a resolution of 768, random flipping, and random adjustments of contrast, hue, brightness, and saturation applied uniformly to all training pairs.

%steps
We employ a progressive training strategy to divide the whole process into three distinct stages. All three stages are executed on 3 NVIDIA GeForce RTX 3090 GPUs with a batch size of 1. The pretrained weights of Stable Diffusion V2.1 are used to initialize the U-net and ControlNet. We employ the AdamW~\cite{loshchilov2017decoupled} optimizer with a fixed learning rate of $3 \times 10^{-4}$ except during the reflection-invariant finetuning phase, where a learning rate of $1 \times 10^{-4}$ is used. To optimize memory usage during the reflection-invariant finetuning in the second phase, we alternate the training of the ControlNet and the upsampling blocks of the U-Net every 100 steps. The durations for the three parts are 2 days, 1 day, and 1 day, respectively. Our method achieves fast inference speed, processing a 768-resolution image in approximately  1 second on a single NVIDIA GeForce RTX 3090 GPU.

We evaluate the performance of our model on three established benchmarks: $Nature$~\cite{li2020single}, $Real$~\cite{zhang2018single}, and $SIR^2$~\cite{wan2017benchmarking}. Additionally, we introduce a new benchmark derived from our DRR dataset to provide a more comprehensive evaluation. Our DRR dataset includes one standard set and another challenging set, each containing 200 image pairs sampled from 20 different scenes. This design further validates the robustness and generalization capabilities of our approach across diverse real-world conditions.

\section{More Qualitative Comparison}
\begin{figure*}[t]
\captionsetup{type=figure}
\centering
\resizebox{\linewidth}{!}{
\includegraphics[width=1.0\textwidth]{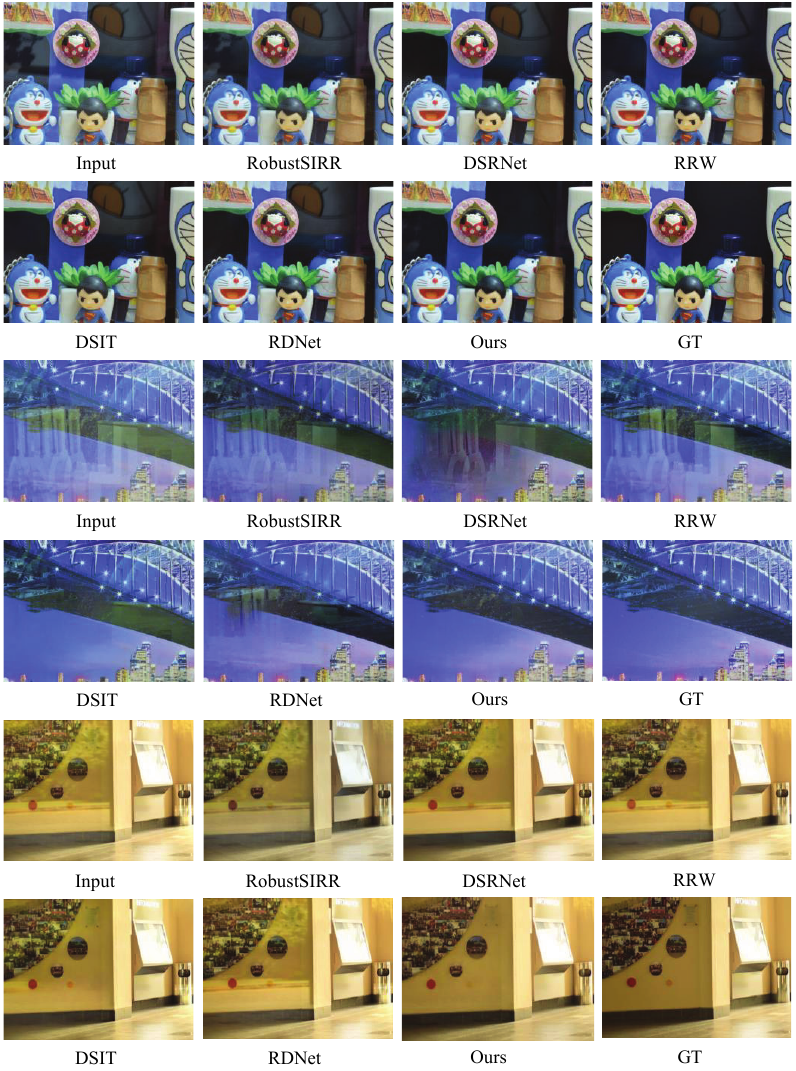}}
\caption {More qualitative comparison of our method with other approaches (not finetuned on our data setting), tested on $SIR^2$~\cite{wan2017benchmarking}.}
\label{fig: compare_sir}
\end{figure*}

\begin{figure*}[t]
\captionsetup{type=figure}
\centering
\resizebox{\linewidth}{!}{
\includegraphics[width=1.0\textwidth]{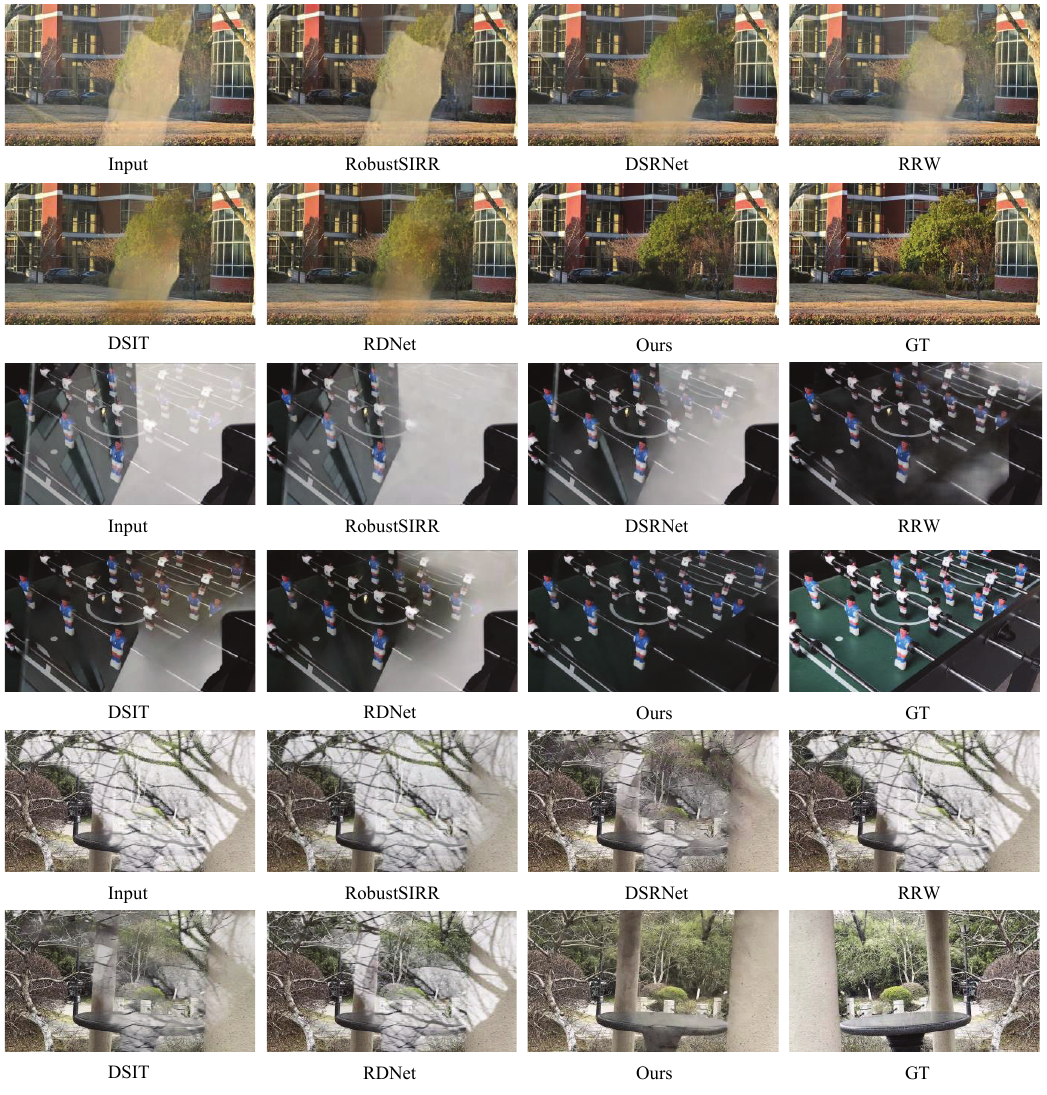}}
\caption {Qualitative comparison of our method with other approaches (not finetuned on our data setting), tested on $DRR$-$S$ and $DRR$-$C$.}
\label{fig: compare_ours}
\end{figure*}

We provide additional qualitative comparisons between our method and other approaches, as illustrated in Fig.~\ref{fig: compare_sir} and Fig.~\ref{fig: compare_ours}, highlighting the superior performance of our comprehensive method.

\section{Dereflection Any Image}
\begin{figure}[t]
\captionsetup{type=figure}
\centering
\resizebox{\linewidth}{!}{
\includegraphics[width=1.0\textwidth]{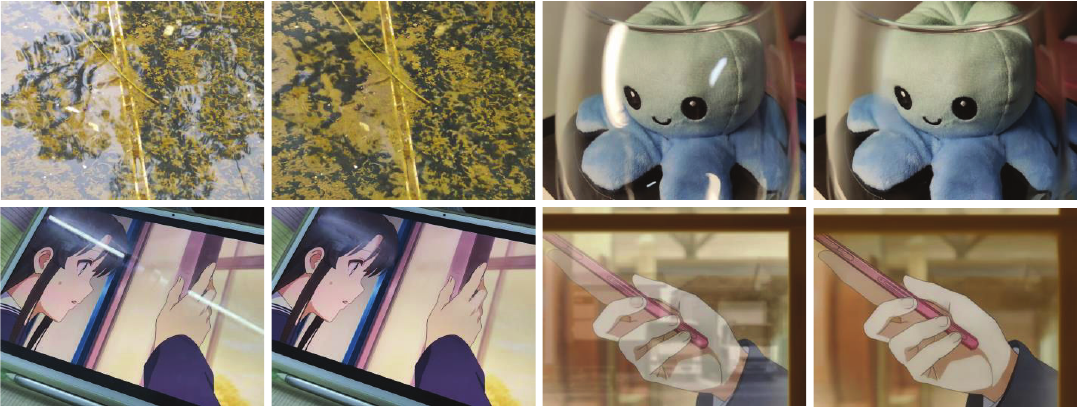}}
\caption {Visualization of the generalization capability across diverse real-world reflection patterns. Our proposed method exhibits robust generalization to diverse reflection types unseen during training, including reflections from water surfaces, glossy plastics, digital displays, and even stylized scenes such as anime.}
\label{fig:de_any_image}
\end{figure}

To further demonstrate the versatility and generalization capability of our proposed method, we conduct additional experiments on challenging real-world images exhibiting various complex reflection patterns, such as reflections on water surfaces, glossy plastics, digital screens, and even stylized anime scenes. Due to the absence of ground-truth transmission layers in these scenarios, our evaluation is qualitative. Visualization results in Fig.~\ref{fig:de_any_image} clearly illustrate the robustness and effectiveness of our approach in handling diverse, uncontrolled reflection conditions.

\section{Downstream application}
\begin{figure*}[h]
\captionsetup{type=figure}
\resizebox{\linewidth}{!}{
\includegraphics[width=1.0\textwidth]{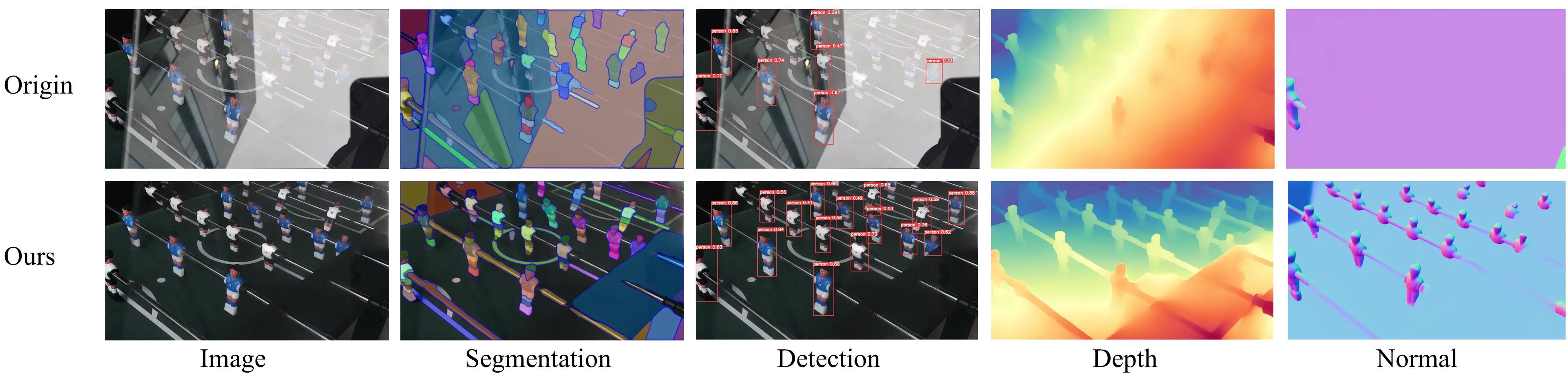}}
\caption {Various downstream tasks, including semantic segmentation, object detection, depth estimation, and normal estimation.}
\label{fig: downstream}
\end{figure*}

Our model can be applied to various downstream tasks, including semantic segmentation, object detection, depth estimation, and normal estimation. The results in Fig.~\ref{fig: downstream} demonstrate that our model can effectively remove reflections and restore the original scene, which is beneficial for subsequent tasks.

\begin{table*}[t]
\centering

\label{tab:old_metrics}
\fontsize{9}{10}\selectfont
\begin{tabular}{lr|cccccc|ccccc}
\toprule
\multirow{2}{*}{Methods} & \multirow{2}{*}{Venue} & \multicolumn{2}{c}{$Nature$ (20)} & \multicolumn{2}{c}{$Real$ (20)} & \multicolumn{2}{c|}{$SIR^{2}$ (500)} & \multicolumn{2}{c}{$DRR$-$S$ (200)} & \multicolumn{2}{c}{$DRR$-$C$ (200)} \\ \cmidrule(l{2pt}r{2pt}){3-4} \cmidrule(l{2pt}r{2pt}){5-6} \cmidrule(l{2pt}r{2pt}){7-8} \cmidrule(l{2pt}r{2pt}){9-10} \cmidrule(l{2pt}r{2pt}){11-12}
 & & PSNR$\uparrow$ & SSIM$\uparrow$ & PSNR$\uparrow$ & SSIM$\uparrow$ & PSNR$\uparrow$ & SSIM$\uparrow$ & PSNR$\uparrow$ & SSIM$\uparrow$ & PSNR$\uparrow$ & SSIM$\uparrow$ \\ \midrule
RobustSIRR & CVPR 2023 & 20.94 & 0.770 & 22.71 & 0.787 & 22.61 & 0.872&  19.68         &    0.756   & 20.24 & 0.692  \\
DSRNet  & ICCV 2023 & 24.86 & 0.823 & 23.31 & 0.791 & 25.65 & 0.919 &  22.33         &    0.846   & 21.93 & 0.820 \\
RRW  & CVPR 2024 & 25.79 & 0.833 & 21.51 & 0.767 & 25.31 & 0.907 &  22.39         &    0.857   & 21.84 & 0.820 \\
L-DiffER & ECCV 2024 & 23.95 & 0.831 & 23.77 & 0.821 & 25.18 & 0.911 & - & - & - & - \\
DSIT  & NeurIPS 2024 & 26.25 & 0.833 & 24.54 & 0.814 & 26.34 & 0.922 &  23.48         &    0.869   & 22.46 & 0.817  \\
RDNet & CVPR 2025 & 25.94 & 0.835 & 25.10 & \underline{0.833} & \underline{26.38} & 0.924 &  23.31         &    0.865   & 22.25 & 0.823  \\ \midrule
RobustSIRR$^\dagger$  & CVPR 2023 & 23.74 & 0.818 & 23.00 & 0.793 & 24.14 & 0.897 &  23.25         &    0.793   & 20.94 & 0.709\\
DSRNet$^\dagger$  & ICCV 2023 & 25.29 & 0.827 & 24.52 & 0.822 & 25.97 & 0.925 &  24.47         &    0.874   & 22.13 & 0.826 \\
RRW$^\dagger$  & CVPR 2024 & \underline{26.76} & \textbf{0.847} & 24.24 & 0.809 & 25.32 & 0.916 &  23.38         &    0.869   & 22.24 & 0.823  \\
DSIT$^\dagger$  & NeurIPS 2024 & 26.30 & 0.833 & \underline{25.18} & 0.822 & 26.20 & \underline{0.927}  &  \underline{26.46}         &    \underline{0.891}  & \underline{23.34} & \underline{0.834} \\
RDNet$^\dagger$  & CVPR 2025 & 26.72 & 0.841 & 24.80 & 0.827 & 26.09 & 0.923  &  24.96         &    0.879  & 22.17 & 0.827 \\ \midrule
Ours & - & \textbf{27.05} & \underline{0.846} & \textbf{25.24} & \textbf{0.840} & \textbf{27.32} & \textbf{0.931} &  \textbf{27.21}         &    \textbf{0.901}   & \textbf{23.83} & \textbf{0.840}  \\ \bottomrule
\end{tabular}
% \vspace{-1em}
\caption{Quantitative comparisons on the existing three reflection benchmarks, and our new dataset DRR, consisting a standard set (DRR-S) and a challenging set (DRR-C). The scores after finetuning on our data setting are labeled as $\dagger$. The best results are in \textbf{bold}, and the second-best results are \underline{underlined}.}
\end{table*}

\section{Limitations}
\begin{figure}[t]
\captionsetup{type=figure}
\resizebox{\linewidth}{!}{
\includegraphics[width=1.0\textwidth]{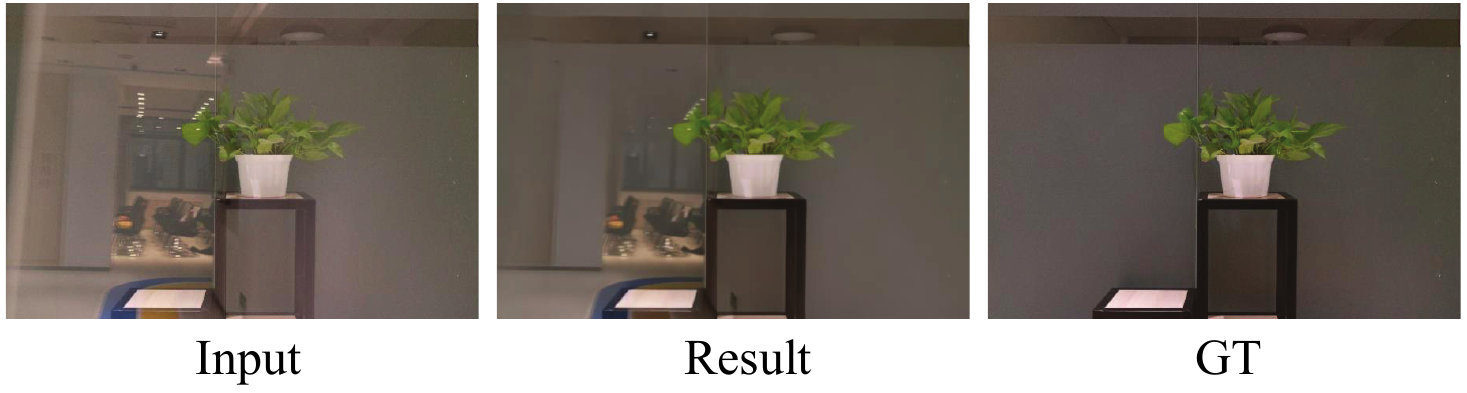}}
\caption {An example of limitation.}
\label{fig: limitation}
\end{figure}

As reflection removal is an ill-posed problem, failure cases still exists. When the reflection and target scene are seamlessly integrated, it is hard to distinguish which one is the target to be preserved. An example is shown in Fig.~\ref{fig: limitation}. To address this limitation, incorporating semantic or user guidance offers a viable solution and a promising direction for future research.

\end{document}